\documentclass[letterpaper, 10 pt, journal, twoside]{IEEEtran}

\usepackage{cite}
\usepackage{graphicx}
\usepackage{booktabs}
\usepackage{amsmath, amssymb}
\usepackage{makecell}
\usepackage{multirow}
\usepackage{threeparttable}
\usepackage{tikz}
\newcommand*{\circled}[1]{\lower.7ex\hbox{\tikz\draw (0pt, 0pt)%
    circle (.5em) node {\makebox[1em][c]{\small #1}};}}

\usepackage{colortbl}
\definecolor{mygray}{gray}{.9}  

\usepackage{float}                  
\usepackage{subfig}                 
\usepackage{overpic}                

\ifCLASSINFOpdf
\else
\fi

\begin{document}

\title{D3PRefiner: A Diffusion-based Denoise Method for 3D Human Pose Refinement}

\author{Danqi Yan, 
        Qing Gao, 
        Yuepeng Qian,
        Xinxing Chen,
        Chenglong Fu,
        and Yuquan Leng
\thanks{This work was supported by the National Natural Science Foundation of China [Grant 52175272]; the Science, Technology and Innovation Commission of Shenzhen Municipality [JCYJ20220530114809021 and KCXFZ20230731093059012, and KCXFZ20230731093401004]; Guangdong Basic and Applied Basic Research Foundation (2022A1515011431); Shenzhen Science and Technology Program (RCBS20210609104516043). \textit{(Corresponding author: Yuquan Leng)}}
\thanks{D. Yan, Y. Qian, X. Chen, C. Fu, and Y. Leng are with the Shenzhen Key Laboratory of Biomimetic Robotics and Intelligent Systems and the Guangdong Provincial Key Laboratory of Human-Augmentation and Rehabilitation Robotics in Universities, Department of Mechanical and Energy Engineering, Southern University of Science and Technology, Shenzhen 518055, China.}
\thanks{Q. Gao is with the School of Electronics and Communication Engineering, Sun Yat-sen University, Shenzhen 518107, China.}
\thanks{This work has been submitted to the IEEE for possible publication. Copyright may be transferred without notice, after which this version may no longer be accessible.}
}

\maketitle

\begin{abstract}
Three-dimensional (3D) human pose estimation using a monocular camera has gained increasing attention due to its ease of implementation and the abundance of data available from daily life.
However, owing to the inherent depth ambiguity in images, the accuracy of existing monocular camera-based 3D pose estimation methods remains unsatisfactory, and the estimated 3D poses usually include much noise. 
By observing the histogram of this noise, we find each dimension of the noise follows a certain distribution,
which indicates the possibility for a neural network to learn the mapping between noisy poses and ground truth poses.
In this work, in order to obtain more accurate 3D poses, a Diffusion-based 3D Pose Refiner (D3PRefiner) is proposed to refine the output of any existing 3D pose estimator.
We first introduce a conditional multivariate Gaussian distribution to model the distribution of noisy 3D poses, using paired 2D poses and noisy 3D poses as conditions to achieve greater accuracy.
Additionally, we leverage the architecture of current diffusion models to convert the distribution of noisy 3D poses into ground truth 3D poses.
To evaluate the effectiveness of the proposed method,
two state-of-the-art sequence-to-sequence 3D pose estimators are used as basic 3D pose estimation models, and the proposed method is evaluated on different types of 2D poses and different lengths of the input sequence.
Experimental results demonstrate the proposed architecture can significantly improve the performance of current sequence-to-sequence 3D pose estimators, 
with a reduction of at least 10.3$\%$ in the mean per joint position error (MPJPE) and at least 11.0$\%$ in the Procrustes MPJPE (P-MPJPE).
\end{abstract}


\IEEEpeerreviewmaketitle

\section{Introduction}

\begin{figure*}[tb]
\centering
\includegraphics[width=0.95\textwidth]{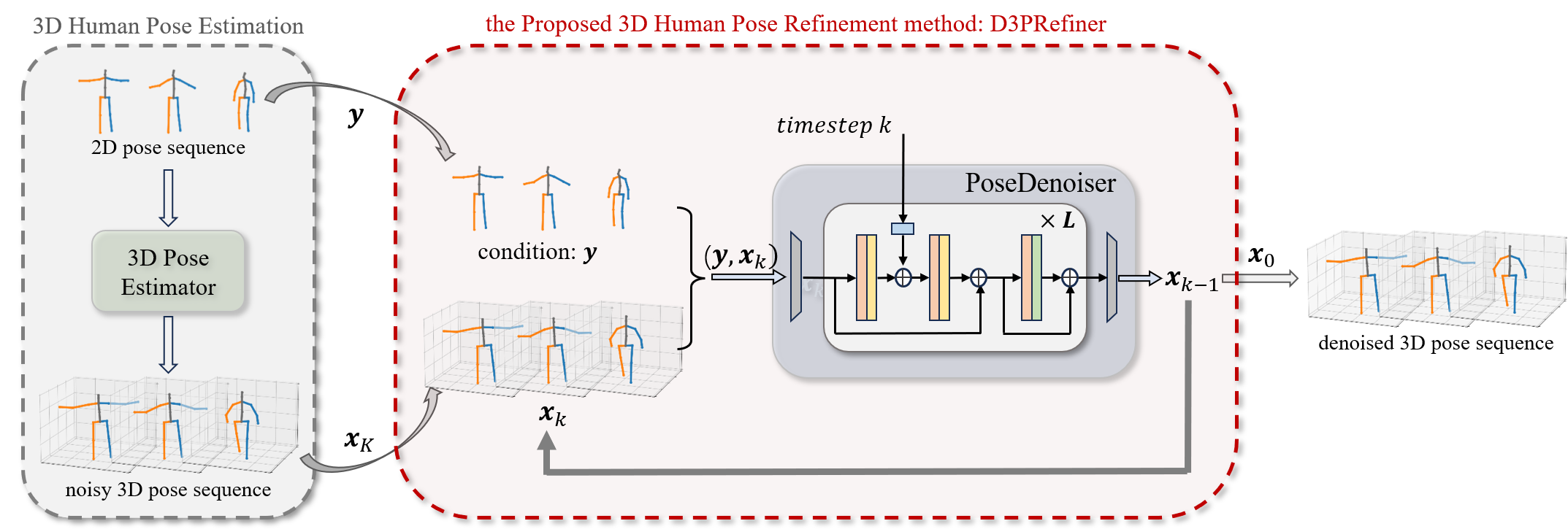}
\caption{
Overview framework of the proposed 3D human pose refinement method.
A 3D human pose estimator takes a sequence of 2D human poses as input and outputs a 3D pose sequence with noise. 
The proposed method uses the 2D pose sequence as a condition to refine the noisy 3D pose sequence in $K$ steps.
}
\label{fig_overview}
\end{figure*}

\IEEEPARstart{O}{ver} the past decades, monocular camera-based 3D human pose estimation has gained more and more attention,
which aims to estimate the 3D location of each joint in a human body from a single image or a sequence of images obtained by a monocular camera. It has shown great potential in many fields, especially for robot learning. State-of-the-art robot learning methods\cite{dong2022passive, hiruma2022deep} aim to allow robots to learn how to behave from an expert's demonstration, and 3D human poses can provide critical information to describe the expert's behavior. However, such learning-based methods are highly dependent on the quantity and quality of the human pose data. Compared to 3D human pose estimation methods using depth cameras \cite{marin2018depth, fang2021rgbd} or inertial measurement units (IMUs) \cite{tang2022imu, bangera2020mems}, monocular camera-based approach can be much cheaper and more convenient, and can utilize the large amount of data from daily life. For example, \cite{sivakumar2022youtube} leveraged a large number of human hands videos from YouTube to acquire 3D poses, which were then utilized for the teleoperation of a robot hand. However, due to inherent depth ambiguity and self-occlusion in an image, accurate 3D human pose estimation using a monocular camera is still challenging.

Traditional methods \cite{agarwal2004silhouettes, onishi2008hog} tend to derive 3D human poses by extracting invariant features, like silhouettes or histograms of oriented gradients, which are easy to operate but vulnerable to external disturbance. Deep learning-based methods \cite{pavllo2019semi,liu2020attention,chen2021anatomy,cai2019exploiting,zhao2023poseformerv2,hossain2018lstm,lin2017recurrent,wang2020motion,hu2021conditionalgraph,zhang2022mixste,tang2023stc,choi2022diffupose} have shown better performance than traditional ones \cite{agarwal2004silhouettes, onishi2008hog}, where various neural networks were designed to learn the mapping between 2D poses and 3D poses.  Some works\cite{martinez2017simple,pavlakos2017coarse,tekin20172dcues,pavlakos2018ordinal} estimated the 3D human pose from a single image, while others \cite{pavllo2019semi,liu2020attention,chen2021anatomy,cai2019exploiting,zhao2023poseformerv2,hossain2018lstm,lin2017recurrent,wang2020motion,hu2021conditionalgraph,zhang2022mixste,tang2023stc,choi2022diffupose} utilized a sequence of images to incorporate additional spatial and temporal information to further alleviate depth ambiguities. Besides, because data obtained from sensors or deep learning models always include noise\cite{yang2023xinxing,rakotosaona2020pointcleannet}, some works\cite{cai2019exploiting, wang2019learning, mei2019learning, zeng2022smoothnet} implemented an additional refinement step to achieve better accuracy: after obtaining the estimated 3D poses, they introduced a second neural network or additional geometric information to refine the former results, which could further improve the accuracy.

In this work, in order to improve the performance of existing monocular-based 3D pose estimation methods, we propose a novel diffusion-based architecture to refine the original output of a 3D pose estimation model. The original outputs of a 3D pose estimation model always include errors, so these outputs can be regarded as noisy 3D poses. The overview of the proposed refinement method D3PRefiner is shown in Fig \ref{fig_overview}, which denoises noisy 3D poses conditioned on 2D poses. The presented method is based on a diffusion architecture, which contains a forward process to gradually add noise on the ground truth 3D poses with several steps and a reverse process to use a neural network to remove the added noise at each step. Unlike general diffusion models that sample noise from a standard Gaussian distribution at each step, we sample noise from the distribution of noisy 3D poses. However, the real distribution of noisy 3D poses is unknown.  To simulate this distribution as accurately as possible, a conditional multivariate Gaussian distribution is introduced, with the paired 2D pose and the original output 3D pose from a 3D pose estimator as conditions. During inference, by iteratively feeding the output of the 3D pose estimator into the reverse process, the denoised 3D poses can be derived. A more detailed explanation is provided in Section \ref{method}.

To the best of our knowledge, the proposed method is the first one to implement diffusion architecture for refining the output of a 3D human pose estimation model, and it also smooths the denoised 3D poses by incorporating the temporal information and reduces the overall offset error by simulating the noisy 3D pose distribution. The main contribution of this work can be summarized by:
1) We propose a novel diffusion-based denoise architecture to refine the output of a $seq2seq$ 3D pose estimation model. 2) We introduce a conditional distribution method to model noisy 3D poses and a novel forward process to reduce uncertainty during inference. 3) The proposed approach can efficiently refine the output of the base 3D pose estimation model and significantly improve the accuracy of state-of-the-art $seq2seq$ methods.
\section{Related Work}
\subsection{Monocular Camera-based 3D Human Pose Estimation}
3D human pose estimation with a monocular camera mainly relies on the development of deep learning techniques, which utilize a large amount of data to build a mapping between input data and output data. Current research can be roughly divided into a one-stage manner\cite{pavlakos2017coarse, tekin20172dcues, lin2017recurrent, pavlakos2018ordinal} and a two-stage manner\cite{martinez2017simple, pavllo2019semi, liu2020attention, chen2021anatomy, cai2019exploiting, zhao2023poseformerv2, hossain2018lstm, wang2020motion, hu2021conditionalgraph, zhang2022mixste, tang2023stc}. For a one-stage manner, the model directly regresses 3D pose from input image data, without intermediate 2D pose representation. The two-stage manner first uses an off-the-shelf 2D pose detector to estimate 2D pose from input RGB images and then uses a 2D-to-3D lifting model to regress 3D pose coordinates. The two-stage manner is widely adopted in recent studies because it can not only benefit from the 2D keypoint detection task but also avoid data lacking problems for paired images and 3D human poses.

Besides, additional temporal information has been exploited to mitigate the inherent depth ambiguities in one image to provide more robust results. For example, inspired by the consistency of bone length across time, \cite{chen2021anatomy} split the whole task into a bone direction prediction and a bone length prediction to derive 3D human poses; \cite{pavllo2019semi} proposed dilated temporal convolutions over 2D human poses to exploit temporal information; Other methods \cite{hossain2018lstm, cai2019exploiting, wang2020motion, hu2021conditionalgraph, zhang2022mixste, tang2023stc, zhao2023poseformerv2} utilized LSTM\cite{lstm} or GCN\cite{gcn} to capture both spatial dependencies and temporal consistencies. Depending on whether the output is a 3D pose of only central frame or a sequence of 3D poses, these methods can be categorized as $seq2frame$ approach \cite{pavllo2019semi, liu2020attention, chen2021anatomy, cai2019exploiting, zhao2023poseformerv2, choi2022diffupose} or $seq2seq$ approach \cite{hossain2018lstm, lin2017recurrent, wang2020motion, hu2021conditionalgraph, zhang2022mixste, tang2023stc}. $seq2frame$ methods usually achieve better performance but cause calculation redundancy, while $seq2seq$ methods can improve the coherence of output 3D poses and avoid unnecessary redundancy. Therefore, to minimize the computation cost and benefit from the temporal information, $seq2seq$ 2D-to-3D lifting models are chosen as basic 3D pose estimation models to regress the 3D human poses, and then the proposed diffusion-based refinement model is implemented in a $seq2seq$ manner.

\begin{figure*}[tb]
\centering
\subfloat[Adding noise sampled from a standard Gaussian distribution]
{\includegraphics[width=0.9\linewidth]{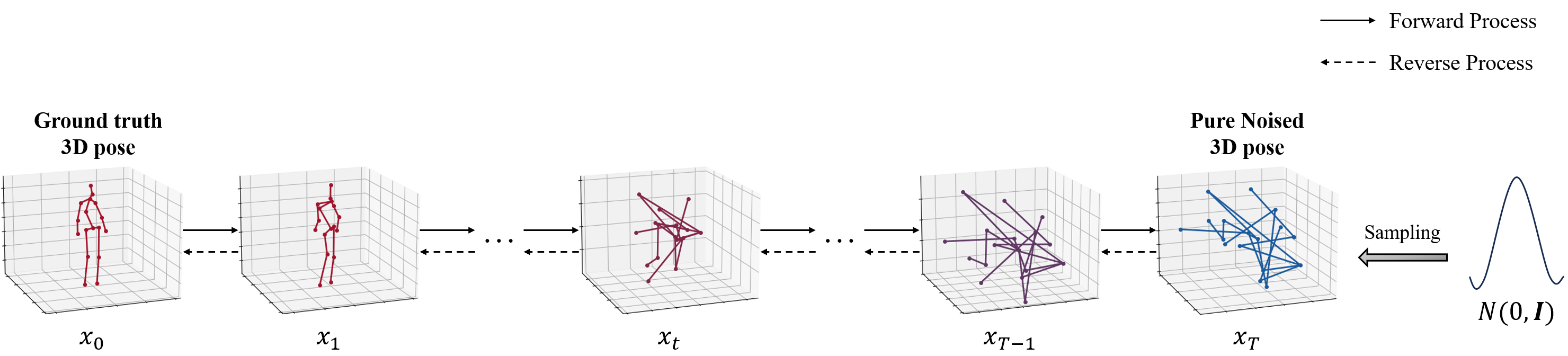}} \\
\subfloat[Adding noise sampled from conditional noisy 3D pose distribution]
{\includegraphics[width=0.9\linewidth]{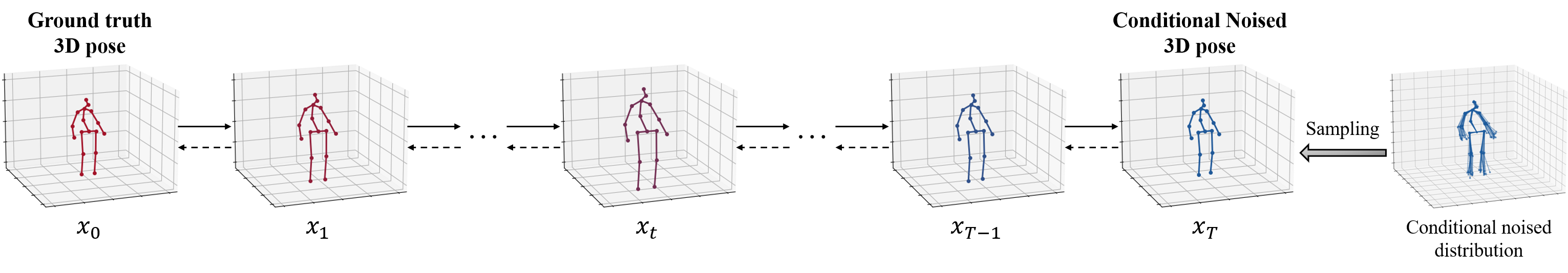}} \\
\caption{
Comparison of noise sampling schemes in the diffusion process. 
Solid black arrows denote the forward process in diffusion models, while dashed black arrows denote the reverse process. (a) visualizes the process of sampling noise from a standard Gaussian distribution $\mathcal{N}(0, \boldsymbol{I})$. The forward process gradually destroys the structure of the ground truth 3D pose until it conforms to $\mathcal{N}(0, \boldsymbol{I})$. (b) visualizes the process of sampling noise from a conditional noisy 3D pose distribution. It progressively transforms the ground truth 3D pose into a reasonable distribution of noisy 3D poses.
}
\label{fig_diff_process_compare}
\end{figure*}

\subsection{Refinement Methods for Human Pose Estimation} \label{related_refine}
Many methods have been explored to refine the estimated 2D or 3D human pose to have more accurate results. To refine 2D human poses, \cite{moon2019posefix} utilized the similar error distributions of state-of-the-art 2D human pose estimation methods as prior information to generate synthetic poses to train their refinement model; \cite{wang2020graph} refined 2D human poses by considering the relationship between keypoints at different locations. For 3D human pose refinement, \cite{wang2019learning} and \cite{mei2019learning} utilized compact representations to avoid illegitimate poses; \cite{zeng2022smoothnet} proposed a temporal-only refinement network by exploiting the natural smoothness characteristics in body movements. Previous works\cite{moon2019posefix, wang2020graph} lacked the temporal information to improve the coherence among a sequence of 3D poses, while \cite{zeng2022smoothnet} and \cite{mei2019learning} incorporated temporal information but ignored the overall offset of output 3D poses. Therefore, we first refine 3D poses in a $seq2seq$ manner to utilize spatial and temporal information. Besides, instead of directly defining the frequency of each pose error (i.e., jitter, inversion, swap, and miss) \cite{moon2019posefix}, a conditional error distribution is used to model the possible noise for each pose, which can define the error distribution more accurately and generate more reasonable noisy poses. By learning the mapping between noisy poses and ground truth poses, the proposed method can effectively reduce the overall offset.

\subsection{Diffusion Models}
Diffusion models are a class of models that can effectively learn data distribution and generate realistic samples from Gaussian noise in $T$ steps. Diffusion models were first proposed by \cite{ho2020ddpm} to generate high-quality images, and by now they have demonstrated their extraordinary capacities on various tasks, like image super-resolution\cite{saharia2022superres} and image-to-image translation\cite{meng2021sdedit}. Some previous studies\cite{choi2022diffupose, shan2023diffusion, gong2023diffpose} also explored its application in 3D human pose estimation. To note that, diffusion models can generate diverse realistic data because $\boldsymbol{x}_{T}$ is a random sample from $\mathcal{N}(0, \boldsymbol{I})$, but 3D human pose estimation task expect the generated results as close to the ground truth as possible, diversity caused by $\boldsymbol{x}_{T}$ reduces accuracy. In order to obtain more robust and accurate 3D poses, \cite{choi2022diffupose, shan2023diffusion, zhou2023diff3dhpe} used multi-hypothesis aggregation strategies: multiple $\boldsymbol{x}_{T}$ were first sampled from $\mathcal{N}(0, \boldsymbol{I})$, and then the average of these corresponding hypotheses was used as the final result. \cite{gong2023diffpose} took the output of another 3D human pose estimation model \cite{zhao2022graformer,zhang2022mixste} as intermediate value $\boldsymbol{x}_{K}$, where $0< K < T$, and then began reverse process from time-step $k$. However, the added noise still followed a standard Gaussian distribution $\mathcal{N}(0, \boldsymbol{I})$, so they could not make sure the intermediate $\boldsymbol{x}_{K}$ had the same distribution as the output estimated 3D poses. In this work, we formulate noisy 3D pose refinement as a reverse process in diffusion models. To generate clean 3D poses from noisy ones, we gradually add noise following the distribution of noisy 3D poses at each time-step, and directly use the output of a 3D pose estimator as $\boldsymbol{x}_{T}$.

\begin{figure*}[tb]
\begin{minipage}[b]{1.0\linewidth}
    \centering
    \subfloat[Training procedure]{\includegraphics[width=0.9\linewidth]{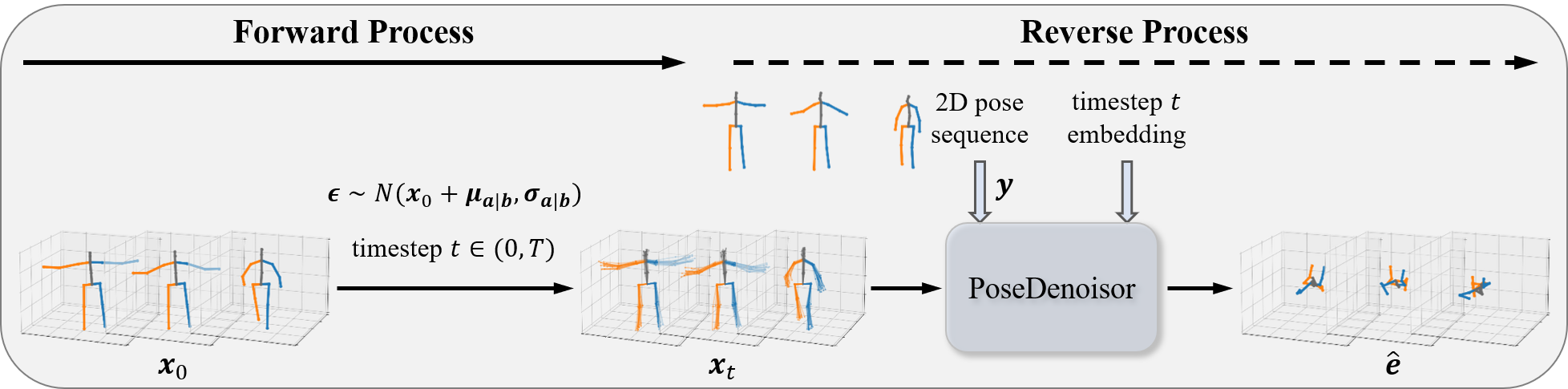}}
\end{minipage}
\begin{minipage}[b]{1.0\linewidth}
    \centering
    \subfloat[Inference procedure]{\includegraphics[width=0.5\linewidth]{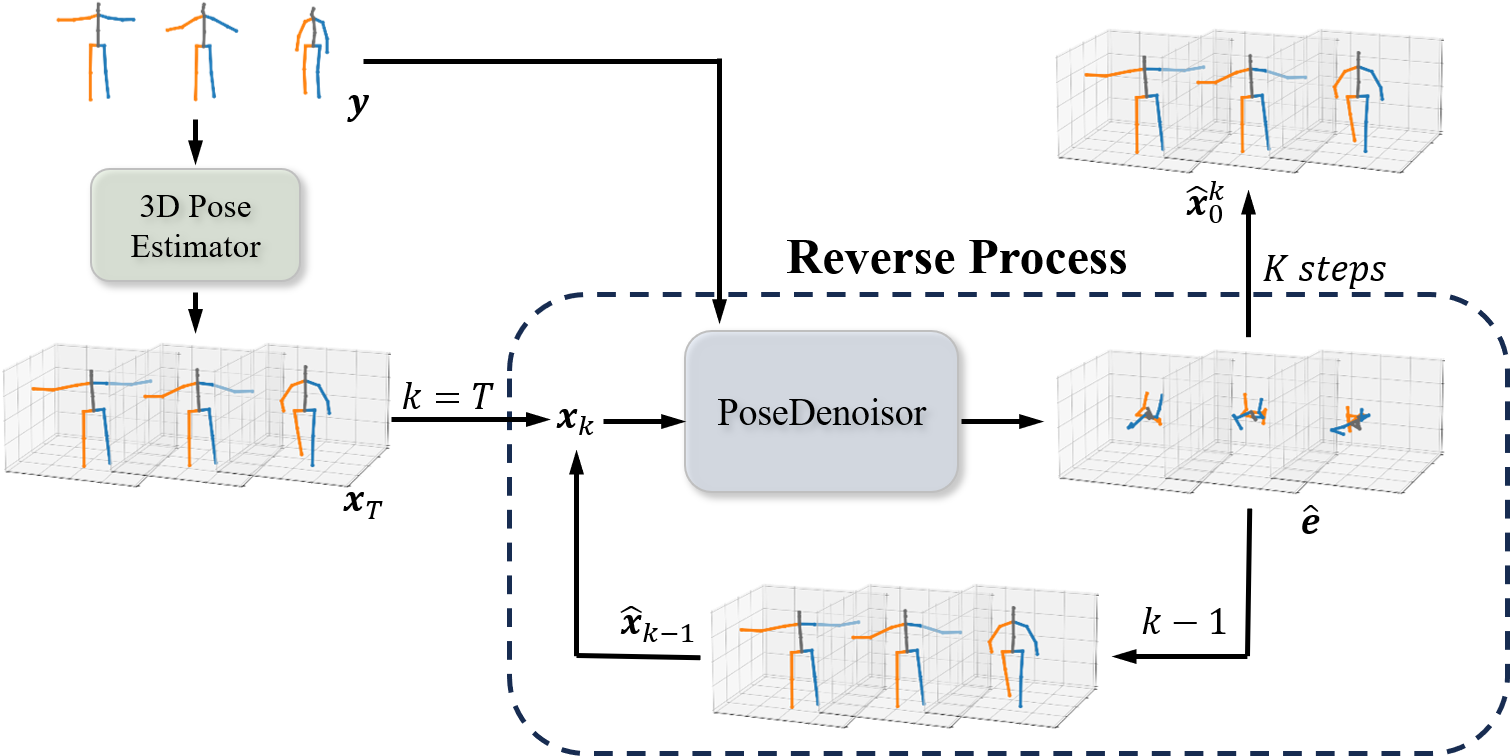}} 
    \hspace{0.03\linewidth}
    \subfloat[Architecture of PoseDenoisor]{\includegraphics[width=0.45\linewidth]{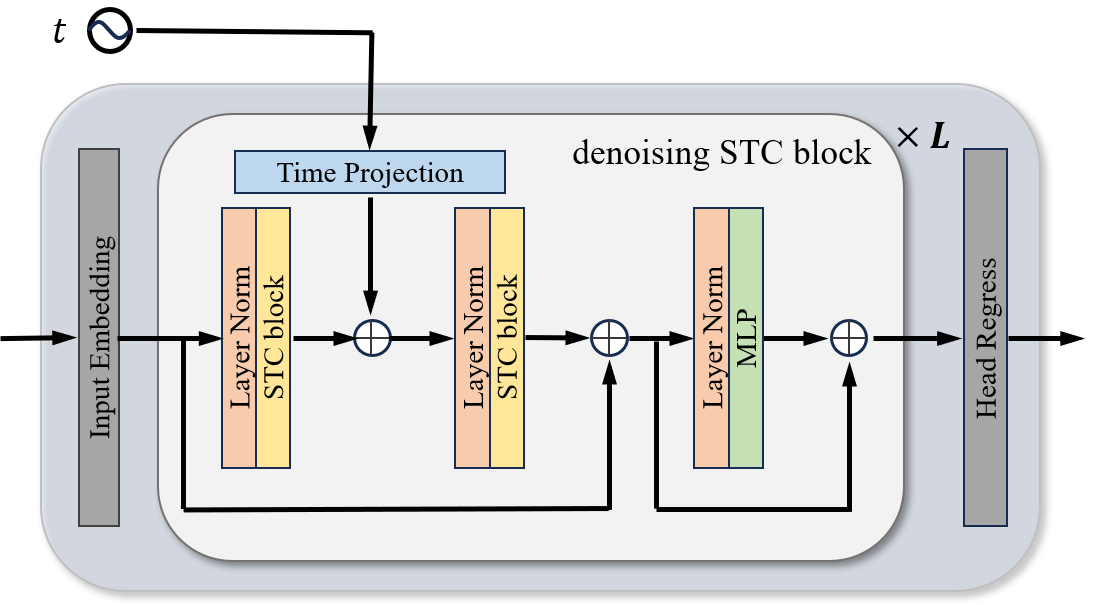}} 
\end{minipage}
\caption{
Overview of the proposed D3PRefiner. (a) shows the training pipeline. In the forward process, noise is sampled from the proposed conditional noisy 3D pose distribution and added to $\boldsymbol{x}_{0}$ with timestep $t$; then in the reverse process, PoseDenoiser is trained to predict error $\boldsymbol{e}$ in it. (b) shows the inference pipeline. The output of a 3D pose estimator is taken as $\boldsymbol{x}_{T}$ and fed into the PoseDenoiser with a 2D pose sequence $\boldsymbol{y}$. After $K$ denoising steps, noise in $\boldsymbol{x}_{T}$ is gradually removed, generating a denoised 3D pose sequence.
}
\label{fig_framework}
\end{figure*}
\section{Method} \label{method}
In this section, a diffusion-based method is proposed to refine the output of a 3D pose estimation model (which can be regarded as noisy 3D poses) with paired 2D poses. A preliminary introduction to the basic idea of diffusion models is first presented, followed by a description of how the distribution of each particular noisy 3D pose can be modeled using a conditional multivariate Gaussian distribution. In addition, the detailed architecture of the proposed D3PRefiner and the corresponding training and inference strategies are illustrated.

\subsection{Background on Diffusion Models}
The proposed method is inspired by diffusion models\cite{ho2020ddpm}, which learns the intricate distribution of input data (like text, images, and so on) by a forward process and a reverse process. In the forward process, Gaussian noise will be gradually added to input data $x_{0}$ in $T$ steps; In the reverse process, a neural network will be trained to remove the added noise and try to restore the original input data $x_{0}$. Once the model has been trained, it can generate expected data from pure Gaussian noise. Fig.\ref{fig_diff_process_compare}(a) shows the general forward and reverse process to generate reasonable 3D poses from Gaussian noise.

To be more specific, assuming input data $\boldsymbol{x}_{0}$ belongs to a real distribution $q(\boldsymbol{x}_{0})$. In the forward process, to gradually add noise to original data $x_{0}$, a parameter $\beta_{t}$ is introduced to control how much noise should be added at the $t$ step, where $0<\beta_{1}<\beta_{2}<\dots<\beta_{T}<1$. This process is based on a Markov chain, so the $t$ step only depends on the previous $t-1$ step, which can be formulated by
\begin{equation}
q(\boldsymbol{x}_{t}|\boldsymbol{x}_{t-1}) 
= 
\mathcal{N}
(
\boldsymbol{x}_{t};
\sqrt{1-\beta_{t}}\boldsymbol{x}_{t-1}, 
\beta_{t}\boldsymbol{I}
)
\label{eq_N_xt_xt-1}
\end{equation}
Eq.(\ref{eq_N_xt_xt-1}) can also be written into
\begin{equation}
\boldsymbol{x}_{t} =
\sqrt{1-\beta_{t}} \boldsymbol{x}_{t-1} + \sqrt{\beta_{t}} \boldsymbol{\epsilon}_{t-1}
\label{eq_xt_xt-1}
\end{equation}
where $\boldsymbol{\epsilon}_{t-1}$ is a little noise at time-step $t-1$ and $\boldsymbol{\epsilon}_{t-1} \sim \mathcal{N}(0, \boldsymbol{I})$. Note that, for $\forall \boldsymbol{\epsilon}_{t}$, there is $\boldsymbol{\epsilon}_{t} \sim \mathcal{N}(0, \boldsymbol{I})$, so all $\boldsymbol{\epsilon}_{t}$ can be replaced by $\boldsymbol{\epsilon}$. Then, if we define $\alpha_{t} = 1-\beta_{t}$, $\overline{\alpha}_{t} = \prod_{t}^{s=0}\alpha_{s}$, $s = 0, 1, ..., t$, $\boldsymbol{x}_{t}$ can be derived by
\begin{equation}
\boldsymbol{x}_{t}
= 
\sqrt{\overline{\alpha}_{t}} \boldsymbol{x}_{0}
+ \sqrt{(1-\overline{\alpha}_{t})} \boldsymbol{\epsilon}
\label{eq_xt_x0}
\end{equation}
In this way, noise at $t$ timestep can be directly added to original data $\boldsymbol{x}_{0}$.

The reverse process is a procedure to remove noise from $\boldsymbol{x}_{t}$. If $q(\boldsymbol{x}_{t-1}|\boldsymbol{x}_{t})$ is known at each timestep, a realistic sample can be gradually generated from $\boldsymbol{x}_{T}$, where $\boldsymbol{x}_{T} \sim \mathcal{N}(0, \boldsymbol{I})$. In the reverse process, a neural network ${\hat{\boldsymbol{\epsilon}}}_{t}=f(\boldsymbol{x}_{t}, t)$ is trained to predict added noise.

\subsection{Conditional Noisy 3D Pose Distribution} \label{sec_conditional_err}
Generally, in each forward process of diffusion models, the added noise is sampled from a standard Gaussian distribution $\mathcal{N}(\boldsymbol{0}, \boldsymbol{I})$. The coefficients $\sqrt{\overline{\alpha}_{t}}$ and $\sqrt{1-\overline{\alpha}_{t}}$ represent the combination ratio of original data $\boldsymbol{x}_{0}$ and noise $\boldsymbol{\epsilon}$. In this way, when time $t$ is big enough, $\sqrt{\overline{\alpha}_{t}} \to 0$ and $\sqrt{1-\overline{\alpha}_{t}} \to 1$, causing $\boldsymbol{x}_{t}$ to converge to $ \boldsymbol{\epsilon}$. This implies $\boldsymbol{x}_{T}$ follow $\mathcal{N}(\boldsymbol{0}, \boldsymbol{I})$, allowing $\boldsymbol{x}_{T}$ to be sampled from pure Gaussian noise to generate realistic data. In a general sense, this process can be viewed as a conversation between two different distributions: one is the distribution of real data $q(\boldsymbol{x}_{0})$, and the other is the distribution of $q(\boldsymbol{\epsilon})$.

Returning to our task, the aim is to refine the output of a 3D pose estimation model to make it closer to the ground truth. The output of a 3D pose estimation model always contains some error, making it different from the ground truth, and this output is called the noisy 3D pose. Hence, 3D pose refinement can be interpreted as a transformation between the distribution of ground truth 3D pose and the distribution of noisy 3D pose. If define the distribution of ground truth 3D pose as $q(\boldsymbol{x}_{0})$ and the distribution of noised 3d pose as $q(\boldsymbol{\epsilon})$, the concept of diffusion models can be directly applied on it, enabling the reverse process to learn how to derive $\boldsymbol{x}_{0}$ from a given noisy 3D pose. This process is illustrated in Fig.\ref{fig_diff_process_compare}(b). The challenge lies in how to define $q(\boldsymbol{\epsilon})$. To address this challenge, a conditional multivariate Gaussian distribution is introduced to model $\boldsymbol{\epsilon}$. We first define $\boldsymbol{\epsilon}$ as the sum of ground truth 3D pose $\boldsymbol{x}_{0} \in \mathbb{R}^{N \times J \times 3}$ and some error $\boldsymbol{e} \in \mathbb{R}^{N \times J \times 3}$:
\begin{equation}
    \boldsymbol{\epsilon} = \boldsymbol{x}_{0} + \boldsymbol{e} 
    \label{eq_gt+noise}
\end{equation} 
By observing the histogram of $\boldsymbol{e}$, we find each of its dimensions follows a Gaussian distribution. To model $q(\boldsymbol{e})$ based on each noisy pose, a joint multivariate Gaussian distribution of 2D pose $\boldsymbol{y} \in \mathbb{R}^{N \times J \times 2}$, output of a 3D pose estimation $\boldsymbol{x}^{noisy} \in \mathbb{R}^{N \times J \times 3}$, and corresponding $\boldsymbol{e} \in \mathbb{R}^{N \times J \times 3}$ is first calculated, and then the concatenation of $\boldsymbol{y}$ and $\boldsymbol{x}^{noisy}$ is used as condition to calculate the distribution of $\boldsymbol{e}$. Here we use $\boldsymbol{a}$ denotes $[\boldsymbol{e}]$ and use $\boldsymbol{b}$ denotes $[\boldsymbol{y}, \boldsymbol{x}^{noisy}]$, the conditional distribution of $\boldsymbol{e}$ can be represented by $ \mathcal{N}(\boldsymbol{\mu}_{a|b}, \boldsymbol{\sigma}_{a|b})$, where
\begin{equation}
\boldsymbol{\mu}_{a|b} 
= 
\boldsymbol{\mu}_{a} + \boldsymbol{\sigma}_{ab} \boldsymbol{\sigma}_{bb}^{-1} (\boldsymbol{x}_{2} - \boldsymbol{\mu}_{b}) 
\label{eq_con_mu}
\end{equation}
\begin{equation}
\boldsymbol{\sigma}_{a|b} 
=
\boldsymbol{\sigma}_{aa} - \boldsymbol{\sigma}_{ab} \boldsymbol{\sigma}_{bb}^{-1} \boldsymbol{\sigma}_{ba}
\label{eq_con_sigma}
\end{equation}

Based on Eq.(\ref{eq_gt+noise}), the distribution of $\boldsymbol{\epsilon}$ could be modeled as $\boldsymbol{\epsilon} \sim \mathcal{N}(\boldsymbol{x}_{0}+\boldsymbol{\mu}_{a|b}, \boldsymbol{\sigma}_{a|b})$. Therefore, we can sample possible noisy 3D pose by given $[\boldsymbol{y}, \boldsymbol{x}^{noisy}]$.

\subsection{Refining Noisy 3D Poses}    
As mentioned above, the architecture of diffusion models can be utilized to refine the output of 2D-to-3D lifting models in order to obtain more accurate results. In the forward process, instead of other diffusion models which sample $\boldsymbol{\epsilon}$ from $\mathcal{N}(0, \boldsymbol{I})$, this work samples $\boldsymbol{\epsilon}$ from $\mathcal{N}(\boldsymbol{x}_{0} + \boldsymbol{\mu}_{a|b}, \boldsymbol{\sigma}_{a|b})$. In the reverse process, a neural network PoseDenoiser is trained $\hat{\boldsymbol{e}} = g(\boldsymbol{y}, \boldsymbol{x}_{t}, t)$ to estimate the error $\boldsymbol{e}$ in noisy 3D poses. Details are shown in the following. 

\subsubsection{PoseDenoiser}
PoseDenoiser is a neural network used in the reverse process of the diffusion architecture, which can be defined as $g(\boldsymbol{y}, \boldsymbol{x}_{t}, t)$, and its detailed architecture is shown in Fig. \ref{fig_framework}(c). The inputs to this neural network are the noisy 3D pose sequence $\boldsymbol{x}_{t} \in \mathbb{R}^{N \times J \times 3}$ at $t$ step, and the corresponding 2D pose sequence $\boldsymbol{y} \in \mathbb{R}^{N \times J \times 2}$, and the embedded timestep $t$. The output of this neural network is the conditional error $\hat{\boldsymbol{e}} \in \mathbb{R}^{N \times J \times 3}$, which is calculated by \ref{sec_conditional_err}. PoseDenoiser is composed of two linear layers and several denoising STC blocks. Each denoising STC block is composed of two STC modules \cite{tang2023stc} and one MLP module with residual connections, where the first STC module is used to catch the temporal and spatial relationship of the input sequence data, and the second one is used to catch the relationship between input data and embedded $t$. The number of stacked denoising STC blocks is $L$. The first linear layer is used to map the input data $(\boldsymbol{x}_{t}, \boldsymbol{y}) \in \mathbb{R}^{N \times J \times 5}$ to a latent embedding $\mathbb{R}^{N \times J \times 256}$, and the second layer is used to map the latent embedding from $\mathbb{R}^{N \times J \times 256}$ to the predicted error $\hat{\boldsymbol{e}} \in \mathbb{R}^{N \times J \times 3}$. The embedded timestep $t$ is first generated for each diffusion step and then projected to each denoising STC block.

\subsubsection{Training}
Training procedure includes both forward process and reverse process, as shown in Fig.\ref{fig_framework}(a). In forward process, the distribution of conditional noisy 3D pose $\mathcal{N}(\boldsymbol{x}_{0}+\boldsymbol{\mu}_{a|b}, \boldsymbol{\sigma}_{a|b})$ is first calculated by given $[\boldsymbol{y}, \boldsymbol{x}^{pred}]$ using Eq.(\ref{eq_con_mu}) and Eq.(\ref{eq_con_sigma}), and then $\boldsymbol{\epsilon}$ is sampled from this distribution where noisy error $\boldsymbol{e}=\boldsymbol{\epsilon}-\boldsymbol{x}_{0}$. After that, a timestep t is randomly sampled from $(0, T]$, and $\boldsymbol{x}_{t}$ is directly calculated by Eq.(\ref{eq_xt_x0}) based on specified $t$, $\boldsymbol{x}_{0}$, and $\boldsymbol{\epsilon}$.
In reverse process, the proposed neural network $\hat{\boldsymbol{e}} = g_{\theta}(\boldsymbol{y}, \boldsymbol{x}_{t}, t)$ is used to estimate noisy error $\boldsymbol{e}$. The loss function is defined as follows:
\begin{equation}
\begin{aligned}
\mathcal{L} 
=& 
\mathbb{E}_{t, \boldsymbol{x}_{0}, \boldsymbol{e}}
\left[ \left\|
\boldsymbol{e} - \hat{\boldsymbol{e}}
\right\|_2 \right] \\
=&
\mathbb{E}_{t, \boldsymbol{x}_{0}, \boldsymbol{e}}
\left[ \left\|
\boldsymbol{e} - g_{\theta}(\boldsymbol{y}, \sqrt{\overline{\alpha}_{t}} \boldsymbol{x}_{0} + (1-\overline{\alpha}_{t}) \boldsymbol{\epsilon}, t)
\right\|_2 \right]
\end{aligned}
\end{equation}

\subsubsection{Inference}
The inference procedure only contains the reverse process, as shown in Fig.\ref{fig_framework}(b). The output of a 2D-to-3D lifting model $\boldsymbol{x}^{noisy}$ is used as $\boldsymbol{x}_{T}$, and the 2D pose sequence $y$ is used as the condition. Then, DDIM\cite{song2020ddim} scheme is implemented to accelerate the reverse process, which means $\boldsymbol{x}_{T}$ is recurrently fed into the neural network $\hat{\boldsymbol{e}} = g_{\theta}(\boldsymbol{y}, \boldsymbol{x}_{k}, k)$ for K times, where $K < T$. Based on Eq.(\ref{eq_gt+noise}) and Eq.(\ref{eq_xt_x0}), we have 
\begin{equation}
\begin{aligned}
\boldsymbol{x}_{k}
=& 
\sqrt{\overline{\alpha}_{k}} \boldsymbol{x}_{0}
+ \sqrt{1-\overline{\alpha}_{k}} \boldsymbol{\epsilon}   \\
=&
\sqrt{\overline{\alpha}_{k}} \boldsymbol{x}_{0}
+ \sqrt{1-\overline{\alpha}_{k}} (\boldsymbol{x}_{0} + \boldsymbol{e})
\end{aligned}
\label{eq_x0_bar_and_xk-1}
\end{equation}
So $\hat{\boldsymbol{x}}_{0}$ and $\boldsymbol{x}_{k-1}$ can be calculated by 
\begin{equation}
\begin{aligned}
&\hat{\boldsymbol{x}}_{0}
=
\frac
{\boldsymbol{x}_{k} - \sqrt{1 - \overline{\alpha}_{k}} \boldsymbol{e}}
{\sqrt{1 - \overline{\alpha}_{k}} + \sqrt{\overline{\alpha}_{k}}} \\
&\boldsymbol{x}_{k-1}
=
(\sqrt{1 - \overline{\alpha}_{k-1}} + \sqrt{\overline{\alpha}_{k-1}}) \hat{\boldsymbol{x}}_{0}
+ \sqrt{1 - \overline{\alpha}_{k-1}} \boldsymbol{e}
\end{aligned}
\end{equation}
\section{Experiments and Results}

\begin{table*}[!t]
\centering
\caption{
Results of applying the proposed refinement method to the state-of-the-art methods. Input 2D pose is estimated by CPN\cite{chen2018cpn}. T denotes the number of frames in each sequence, and the red color indicates the percentage drop by introducing the proposed refinement method.} \label{table_cpn}
\setlength{\tabcolsep}{1.5mm}{
\begin{tabular}{l|llllllllllllllll} 
  \toprule
  MPJPE (mm) $\downarrow$ & Dir. & Disc. & Eat & Greet & Phone & Photo & Pose & Pur. & Sit & SitD. & Smoke & Wait & WalkD. & Walk & WalkT. & Avg.  \\
  \midrule
   MixSTE (T=81) & 39.8 & 43.0 & 38.6 & 40.1 & 43.4 & 50.6 & 40.6 & 41.4 & 52.2 & 56.7 & 43.8 & 40.8 & 43.9 & 29.4 & 30.3 & 42.4 \\
   \rowcolor{mygray} MixSTE+ours (T=81) & 34.0 & 37.3 & 34.3 & 35.9 & 38.8 & 45.5 & 36.0 & 36.7 & 47.4 & 53.3 & 38.7 & 36.8 & 39.3 & 24.9 & 26.0 & 37.7$_{\textcolor{red}{\downarrow11.1\%}}$ \\
   MixSTE (T=243) & 37.6 & 40.9 & 37.3 & 39.7 & 42.3 & 49.9 & 40.1 & 39.8 & 51.7 & 55.0 & 42.1 & 39.8 & 41.0 & 27.9 & 27.9 & 40.9 \\
   \rowcolor{mygray} MixSTE+ours (T=243) & 33.2 & 36.4 & 33.2 & 34.8 & 37.4 & 46.0 & 35.1 & 36.9 & 47.0 & 52.2 & 37.6 & 35.3 & 37.2 & 23.6 & 24.5 & 36.7$_{\textcolor{red}{\downarrow10.3\%}}$ \\
   \midrule
   STCFormer (T=81) & 40.6 & 43.0 & 38.3 & 40.2 & 43.5 & 52.6 & 40.3 & 40.1 & 51.8 & 57.7 & 42.8 & 39.8 & 42.3 & 28.0 & 29.5 & 42.0 \\
   \rowcolor{mygray} STCFormer+ours (T=81) & 34.1 & 37.0 & 33.6 & 34.3 & 37.8 & 47.9 & 35.3 & 35.1 & 48.3 & 53.5 & 38.0 & 34.9 & 37.4 & 23.4 & 24.9 & 37.0$_{\textcolor{red}{\downarrow11.9\%}}$ \\
   STCFormer (T=243) & 39.6 & 41.6 & 37.4 & 38.8 & 43.1 & 51.1 & 39.1 & 39.7 & 51.4 & 57.4 & 41.8 & 38.5 & 40.7 & 27.1 & 28.6 & 41.0 \\
   \rowcolor{mygray} STCFormer+ours (T=243) & 34.4 & 36.8 & 33.3 & 34.8 & 37.2 & 46.4 & 34.4 & 33.8 & 48.1 & 53.5 & 37.9 & 35.0 & 36.9 & 23.1 & 23.8 & 36.6$_{\textcolor{red}{\downarrow10.7\%}}$ \\
  \bottomrule
  \toprule
  P-MPJPE (mm) $\downarrow$ & Dir. & Disc. & Eat & Greet & Phone & Photo & Pose & Pur. & Sit & SitD. & Smoke & Wait & WalkD. & Walk & WalkT. & Avg.  \\
  \midrule
   MixSTE (T=81) & 32.0 & 34.2 & 31.7 & 33.7 & 34.4 & 39.2 & 30.2 & 31.8 & 42.9 & 46.9 & 35.5 & 32.0 & 34.4 & 23.6 & 25.2 & 33.9 \\
   \rowcolor{mygray} MixSTE+ours (T=81) & 26.5 & 28.8 & 26.5 & 28.5 & 29.4 & 34.1 & 27.2 & 26.7 & 37.8 & 42.3 & 30.7 & 27.2 & 30.0 & 19.0 & 20.5 & 29.0$_{\textcolor{red}{\downarrow14.5\%}}$ \\
   MixSTE (T=243) & 30.8 & 33.1 & 30.3 & 31.8 & 33.1 & 39.1 & 31.1 & 30.5 & 42.5 & 44.5 & 34.0 & 30.8 & 32.7 & 22.1 & 22.9 & 32.6 \\
   \rowcolor{mygray} MixSTE+ours (T=243) & 26.5 & 28.5 & 26.6 & 27.7 & 28.6 & 34.7 & 26.8 & 26.9 & 38.4 & 41.2 & 29.8 & 26.7 & 28.8 & 18.1 & 19.4 & 28.6$_{\textcolor{red}{\downarrow12.3\%}}$ \\
   \midrule
   STCFormer (T=81) & 30.4 & 33.8 & 31.1 & 31.7 & 33.5 & 39.5 & 30.8 & 30.0 & 41.8 & 45.8 & 34.3 & 30.1 & 32.8 & 21.9 & 23.4 & 32.7 \\
   \rowcolor{mygray} STCFormer+ours (T=81) & 26.4 & 28.8 & 26.7 & 27.4 & 28.9 & 35.2 & 26.9 & 25.8 & 38.6 & 42.3 & 30.5 & 26.3 & 29.0 & 18.1 & 19.7 & 28.7$_{\textcolor{red}{\downarrow12.2\%}}$ \\
   STCFormer (T=243) & 29.5 & 33.2 & 30.6 & 31.0 & 33.0 & 38.0 & 30.4 & 29.4 & 41.8 & 45.2 & 33.6 & 29.5 & 31.6 & 21.3 & 22.6 & 32.0 \\
   \rowcolor{mygray} STCFormer+ours (T=243) & 25.9 & 28.8 & 26.9 & 27.6 & 28.2 & 34.6 & 26.2 & 25.8 & 39.1 & 41.9 & 30.0 & 26.4 & 28.8 & 18.2 & 18.8 & 28.5$_{\textcolor{red}{\downarrow11.0\%}}$ \\
  \bottomrule
\end{tabular}
}
\end{table*}

\begin{table*}[!t]
\centering
\caption{
Results of applying the proposed refinement method to the state-of-the-art methods. The methods take the ground-truth 2D pose as input. T denotes the number of frames in each sequence, and the red color indicates the percentage drop by introducing the proposed refinement method.}   \label{table_gt}
\setlength{\tabcolsep}{1.5mm}{
\begin{tabular}{l|llllllllllllllll} 
\toprule
  MPJPE (mm) $\downarrow$ & Dir. & Disc. & Eat & Greet & Phone & Photo & Pose & Pur. & Sit & SitD. & Smoke & Wait & WalkD. & Walk & WalkT. & Avg.  \\
  \midrule
   MixSTE (T=81)$^\dag$ & 25.6 & 27.8 & 24.5 & 25.7 & 24.9 & 29.9 & 28.6 & 27.4 & 29.9 & 29.0 & 26.1 & 25.0 & 25.2 & 18.7 & 19.9 & 25.9 \\
   \rowcolor{mygray} MixSTE+ours (T=81)$^\dag$ & 16.7 & 18.6 & 17.9 & 16.8 & 19.3 & 24.7 & 19.8 & 19.5 & 26.1 & 26.5 & 20.3 & 19.5 & 19.5 & 12.7 & 13.2 & 19.4$_{\textcolor{red}{\downarrow25.1\%}}$ \\
   MixSTE (T=243)$^\dag$ & 21.6 & 22.0 & 20.4 & 21.0 & 20.8 & 24.3 & 24.7 & 21.9 & 26.9 & 24.9 & 21.2 & 21.5 & 20.8 & 14.7 & 15.7 & 21.6 \\
   \rowcolor{mygray} MixSTE+ours (T=243)$^\dag$ & 14.6 & 16.4 & 18.0 & 15.3 & 17.7 & 20.9 & 17.5 & 19.3 & 24.2 & 24.1 & 17.4 & 17.0 & 16.5 & 10.9 & 11.7 & 17.4$_{\textcolor{red}{\downarrow19.4\%}}$ \\
   \midrule
   STCFormer (T=81)$^\dag$ & 26.2 & 26.5 & 23.4 & 24.6 & 25.0 & 28.6 & 28.3 & 24.6 & 30.9 & 33.7 & 25.7 & 25.3 & 24.6 & 18.6 & 19.7 & 25.7 \\
   \rowcolor{mygray} STCFormer+ours (T=81)$^\dag$ & 14.3 & 16.2 & 16.2 & 14.2 & 17.8 & 21.3 & 17.2 & 16.1 & 23.1 & 24.3 & 17.4 & 15.7 & 15.3 & 9.5 & 10.7 & 16.6$_{\textcolor{red}{\downarrow35.4\%}}$ \\
   STCFormer (T=243)$^\dag$ & 21.4 & 22.6 & 21.0 & 21.3 & 23.8 & 26.0 & 24.2 & 20.0 & 28.9 & 28.0 & 22.3 & 21.4 & 20.1 & 14.2 & 15.0 & 22.0 \\
   \rowcolor{mygray} STCFormer+ours (T=243)$^\dag$ & 13.2 & 15.3 & 14.7 & 12.9 & 16.5 & 19.7 & 16.5 & 15.3 & 21.1 & 21.5 & 16.6 & 14.0 & 14.6 & 8.5 & 9.2 & 15.3$_{\textcolor{red}{\downarrow30.5\%}}$ \\
  \bottomrule
  \end{tabular}
  }
\end{table*}

\subsection{Dataset and Evaluation Metrics}
We evaluate the proposed method on Human3.6M\cite{ionescu2013h36m}, which is the most widely used dataset for 3D human pose estimation. Human3.6M consists of 3.6 million human poses and corresponding images, including 11 subjects and 15 actions. Following the previous work\cite{pavllo2019semi, zhang2022mixste, tang2023stc}, we adopt 17 joints skeleton and use 5 subjects (S1, S5, S6, S7, S8) to train, 2 subjects (S9 and S11) to test.

As for evaluation metrics, we report the two most commonly used protocols for Human3.6M. Protocol 1 is the mean per joint position error (MPJPE), which calculates the mean Euclidean distance between the predicted poses and the ground truth poses. Protocol 2 is the Procrustes MPJPE (P-MPJPE), which first aligns the predicted poses to the ground truth poses via a rigid transformation and then calculates the MPJPE between aligned predicted poses and ground truth poses. For these two protocols, the lower the value, the higher the accuracy.

\subsection{Implement Details}
The proposed method is implemented with Pytorch\cite{paszke2019pytorch}, a public deep-learning platform. In the experiments, to be consistent with previous works\cite{zhang2022mixste, tang2023stc}, we analyze the performance on both the ground truth 2D pose provided by the dataset and the estimated 2D pose provided by a pre-trained CPN detector\cite{chen2018cpn}, and compare accuracy with and without the proposed method on specific sequence lengths (T=81 and T=243). We also follow the common data preprocessing methods \cite{zhang2022mixste, tang2023stc} to normalize the 2D pose and reduce the root position of the 3D pose.

As for the diffusion parameters, the maximum time-step $T$ is set to 50, and the cosine noise scheduler\cite{nichol2021cosine} is used with the offset $s=0.008$. During the inference, the acceleration technique DDIM\cite{song2020ddim} is implemented to accelerate this procedure and the number of iterative reverse processes $K$ is set to 2. For the present PoseDenoiser, the number of DiffDenoisor Blocks $L$ is set to 4, and the output $\hat{\boldsymbol{e}}$ is normalized by its average value. The Adam\cite{kingma2014adam} optimizer is adopted with the momentum parameters $\beta_{1}, \beta_{2}=0.9, 0.999$ and the initial learning rate is set as 1e-3 with a shrink factor of 0.96 after each epoch. The whole architecture is trained and tested on a single NVIDIA RTX 3090 GPU.

\subsection{Refinement Results on State-of-the-art Methods}
To demonstrate the capacity of the proposed method to refine the output of an arbitrary 3D pose estimation model, we choose two state-of-the-art 2D-to-3D lifting models as the basic 3D pose estimation models (MixSTE\cite{zhang2022mixste} and STCFormer\cite{tang2023stc}) and then compare the difference in accuracy with and without the proposed D3PRefiner. This work is in a $seq2seq$ manner and all the experiments follow the original experiments of MixSTE\cite{zhang2022mixste} and STCFormer\cite{tang2023stc} to evaluate on both 81 frames and 243 frames. For the experiment based on STCFormer, we follow \cite{tang2023stc} to use an additional Temporal Downsampling Strategy (TDS) \cite{shan2022tds} to enlarge the receptive field in the temporal domain and avoid an increase in the number of parameters and computational complexity. To maintain consistency with their pre-trained models, TDS is set to 3 for 81 frames and 2 for 243 frames. For the experiment based on MixSTE, we follow \cite{zhang2022mixste} and do not use this strategy.

To demonstrate the effectiveness of D3PRefiner, we compare the value of MPJPE and P-MPJPE of all 15 actions on test subjects (S9 and S11). Table \ref{table_cpn} and Table \ref{table_gt} summarize the results based on 2D poses provided by the CPN detector and the ground truth 2D poses, respectively. The results show that the proposed method significantly improves the accuracy of the original 3D human pose estimation model in all the actions and all the evaluation metrics: for 2D poses obtained by the CPN detector, D3PRefiner reduces both MPJPE and P-MPJPE by an average of 4.5 $mm$ for MixSTE, and reduces MPJPE and P-MPJPE by an average of 4.7 $mm$ and 3.8 $mm$ for STCCFormer respectively; for the ground truth 2D poses, D3PRefiner reduces MPJPE by an average of 5.4 $mm$ and 7.9 $mm$ for MixSTE and STCCFormer, respectively.

\begin{figure}[t]
\centering
\includegraphics[width=0.45\textwidth]{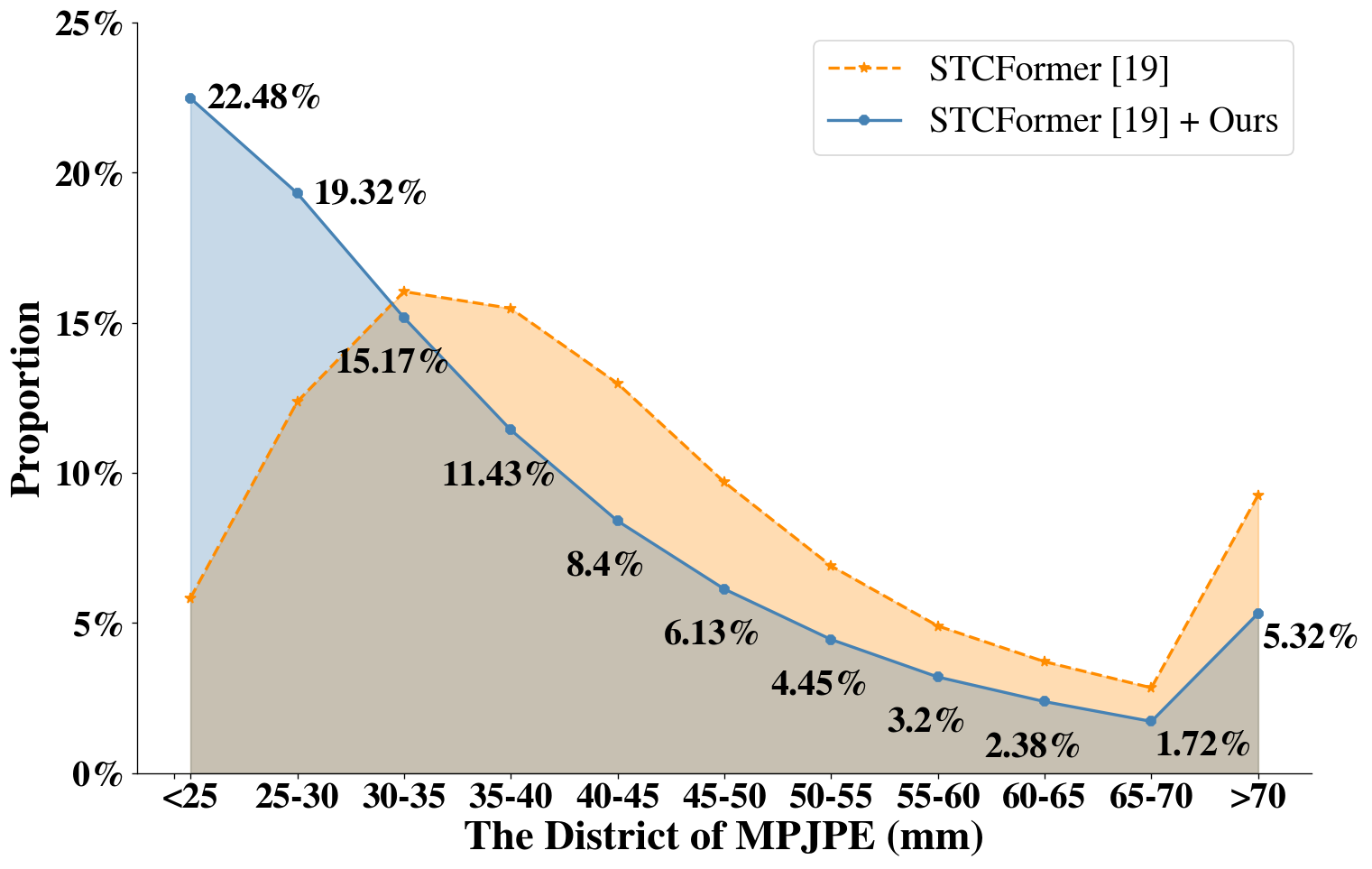}
\caption{MPJPE distribution comparison on Human3.6M}
\label{fig_err_distribution}
\end{figure}

Furthermore, we also compare the error distribution with and without the proposed refinement method. In this experiment, STCFormer is used as the basic 3D pose estimation model, the input sequence contains 81 frames, and 2D poses are provided by the CPN detector. Compared to the basic STCFormer, D3PRefiner significantly increases the proportion for MPJPE $<$ 30 $mm$ and reduces the proportion for MPJPE $>$ 35 $mm$, as shown in Fig.\ref{fig_err_distribution}.

\subsection{Ablation Study}
To evaluate the impact of each component of the proposed method, We conduct several ablation experiments on the Human3.6M dataset and use 2D poses provided by the CPN detector. MixSTE is used as the basic 3D pose estimation model.

\subsubsection{Impact of main components}
D3PRefiner includes two main components: a conditional noisy 3D pose distribution (CND) and a diffusion-based architecture. To demonstrate the effectiveness of the CND, we directly use the output of a 3D pose estimator as noisy 3D poses; to demonstrate the effectiveness of the diffusion-based model, we use a similar architecture without diffusion settings. Table \ref{table_abla_component} shows the experimental results, and the proposed method with both the CND and the diffusion-based architecture achieves the best result. Because the CND can help D3PRefiner learn mappings from noisy 3D distributions, a diffusion-based architecture can provide a more stable and robust learning strategy.

\subsubsection{Impact of the depth of PoseDenoiser}
Table \ref{table_abla_depth} shows how different depths of PoseDenoiser impact the value of MPJPE. Following \cite{tang2023stc} and \cite{zhang2022mixste}, these experiments are conducted under input sequence length T=27. Floating Point Operations (FLOPs) is a commonly used metric to calculate the computational complexity of deep learning models, with lower value indicating lower computational complexity. The results at different depths do not vary much because the output of PoseDenoiser is the normalization error, and this value decreases significantly after denormalization. Based on the results, we choose depth=8 for PoseDenoiser to have a balance between accuracy and computational complexity. 

\subsection{Qualitative Results}
\begin{figure}[!t]
\centering
\includegraphics[width=0.46\textwidth]{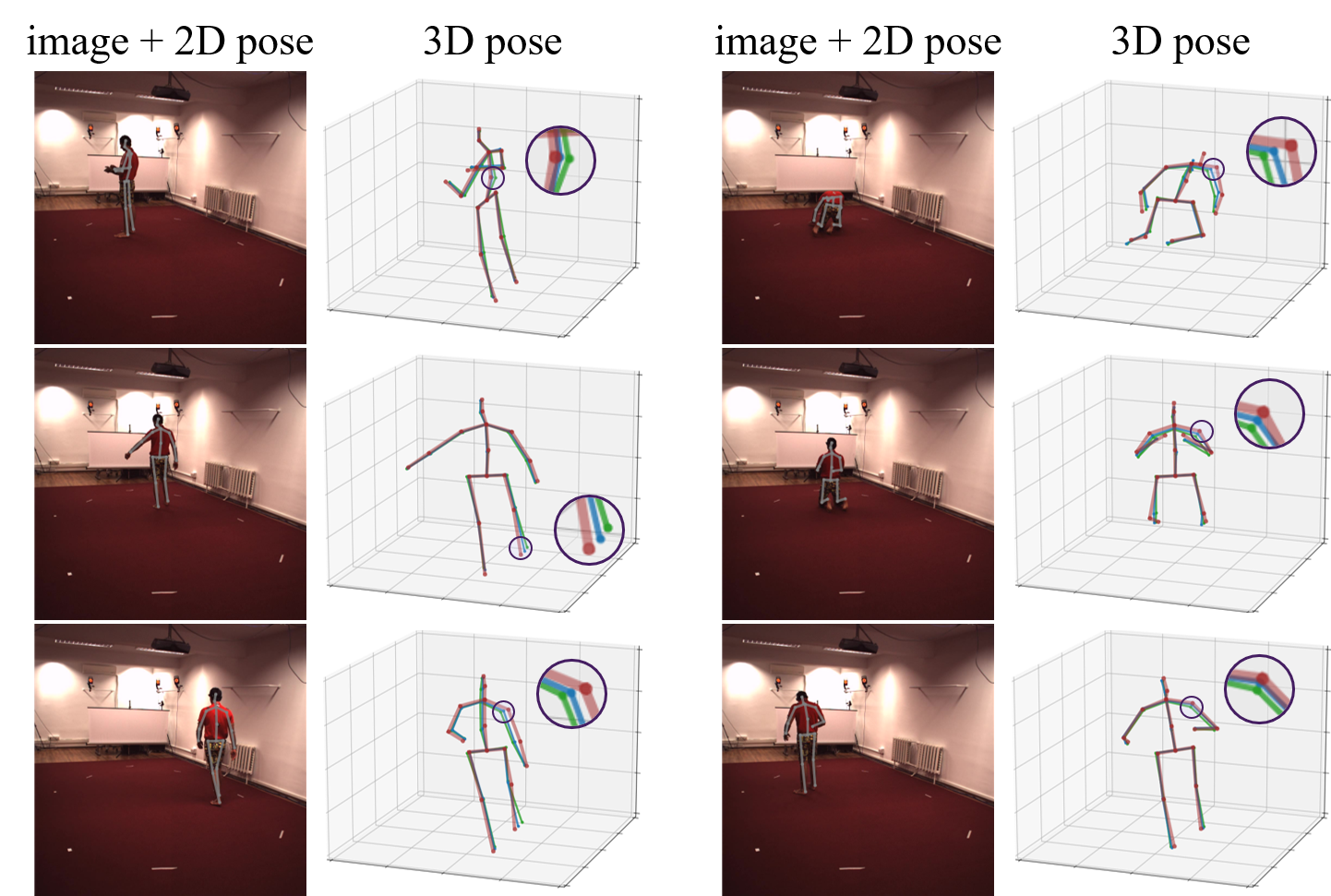}
\caption{Qualitative Results. Red, green, and blue skeletons represent the ground truth 3D pose, the estimated 3D pose and the refined 3D pose by the proposed D3PRefiner, respectively.}
\label{fig_qualitative_results}
\end{figure}

As shown in Fig.\ref{fig_qualitative_results}, we further evaluate the visual result of estimated 3D poses and the ground truth values in Human3.6M. The results shows that our proposed method can further refine the original output of a 3D pose estimation model and obtain more accurate results.

\begin{table}[!t]
\centering
\caption{Ablation study for main components.}
\begin{tabular}{l|lll} 
\toprule
   & MPJPE (mm) $\downarrow$ & P-MPJPE (mm) $\downarrow$ \\
  \midrule
   base   & \textbf{40.0} & \textbf{30.7} \\
   wo. CND   & 46.1 & 35.5 \\
   wo. diff & 40.3 & 30.9 \\
  \bottomrule
  \end{tabular}
  \label{table_abla_component}
\end{table}

\begin{table}[!t]
\centering
\caption{Ablation study for depth of PoseDenoisor.}
\begin{tabular}{l|ll|l} 
\toprule
  depth & FLOPs (M) & Input Length (T) & MPJPE (mm) $\downarrow$ $\downarrow$ \\
  \midrule
   4 & 7.2M & 27 & 40.3 \\
   6 & 10.5M & 27 & 40.2 \\
   8 & 13.9M & 27 & \textbf{40.0} \\
   10 & 17.2M & 27 & \textbf{40.0} \\
  \bottomrule
  \end{tabular}
  \label{table_abla_depth}
\end{table} 

\section{Conclusion}
This work presents a novel seq2seq diffusion-based framework for 3D pose refinement. The present D3PRefiner first simulates the distribution of noisy 3D poses conditioned on paired 2D poses and estimated 3D poses from an arbitrary 3D pose estimation model. By integrating this conditional distribution in the forward diffusion process, PoseDenoiser gradually learns the mapping between the distribution of ground truth 3D poses and noisy 3D poses, which can then recurrently shed noise in the noisy 3D poses. Experimental results demonstrate the proposed method significantly improves the accuracy of state-of-the-art 3D pose estimation models. In the future, we expect to explore more powerful diffusion models and generalize the proposed refinement method to more applications.

\bibliographystyle{IEEEtran}
\bibliography{Reference/refer_ral_learning, Reference/refer_my_3d_pose, Reference/refer_my_bg, Reference/refer_my_refine, Reference/refer_my_diff}

\begin{thebibliography}{10}
\providecommand{\url}[1]{#1}
\csname url@samestyle\endcsname
\providecommand{\newblock}{\relax}
\providecommand{\bibinfo}[2]{#2}
\providecommand{\BIBentrySTDinterwordspacing}{\spaceskip=0pt\relax}
\providecommand{\BIBentryALTinterwordstretchfactor}{4}
\providecommand{\BIBentryALTinterwordspacing}{\spaceskip=\fontdimen2\font plus
\BIBentryALTinterwordstretchfactor\fontdimen3\font minus \fontdimen4\font\relax}
\providecommand{\BIBforeignlanguage}[2]{{%
\expandafter\ifx\csname l@#1\endcsname\relax
\typeout{** WARNING: IEEEtran.bst: No hyphenation pattern has been}%
\typeout{** loaded for the language `#1'. Using the pattern for}%
\typeout{** the default language instead.}%
\else
\language=\csname l@#1\endcsname
\fi
#2}}
\providecommand{\BIBdecl}{\relax}
\BIBdecl

\bibitem{dong2022passive}
Z.~Dong, Z.~Li, Y.~Yan, S.~Calinon, and F.~Chen, ``Passive bimanual skills learning from demonstration with motion graph attention networks,'' \emph{IEEE Robotics and Automation Letters}, vol.~7, no.~2, pp. 4917--4923, 2022.

\bibitem{hiruma2022deep}
H.~Hiruma, H.~Ito, H.~Mori, and T.~Ogata, ``Deep active visual attention for real-time robot motion generation: Emergence of tool-body assimilation and adaptive tool-use,'' \emph{IEEE Robotics and Automation Letters}, vol.~7, no.~3, pp. 8550--8557, 2022.

\bibitem{marin2018depth}
M.~J. Marin-Jimenez, F.~J. Romero-Ramirez, R.~Munoz-Salinas, and R.~Medina-Carnicer, ``3d human pose estimation from depth maps using a deep combination of poses,'' \emph{Journal of Visual Communication and Image Representation}, vol.~55, pp. 627--639, 2018.

\bibitem{fang2021rgbd}
Z.~Fang, A.~Wang, C.~Bu, and C.~Liu, ``3d human pose estimation using rgbd camera,'' in \emph{2021 IEEE International Conference on Computer Science, Electronic Information Engineering and Intelligent Control Technology (CEI)}.\hskip 1em plus 0.5em minus 0.4em\relax IEEE, 2021, pp. 582--587.

\bibitem{tang2022imu}
J.~Tang, H.~Luo, W.~Chen, P.~K.-Y. Wong, and J.~C. Cheng, ``Imu-based full-body pose estimation for construction machines using kinematics modeling,'' \emph{Automation in Construction}, vol. 138, p. 104217, 2022.

\bibitem{bangera2020mems}
S.~S. Bangera, T.~Shiyana, G.~Srinidhi, Y.~R. Vasani, and M.~Sukesh~Rao, ``Mems-based imu for pose estimation,'' in \emph{Advances in Communication, Signal Processing, VLSI, and Embedded Systems: Select Proceedings of VSPICE 2019}.\hskip 1em plus 0.5em minus 0.4em\relax Springer, 2020, pp. 1--14.

\bibitem{sivakumar2022youtube}
A.~Sivakumar, K.~Shaw, and D.~Pathak, ``Robotic telekinesis: Learning a robotic hand imitator by watching humans on youtube,'' \emph{arXiv preprint arXiv:2202.10448}, 2022.

\bibitem{agarwal2004silhouettes}
A.~Agarwal and B.~Triggs, ``3d human pose from silhouettes by relevance vector regression,'' in \emph{Proceedings of the 2004 IEEE Computer Society Conference on Computer Vision and Pattern Recognition, 2004. CVPR 2004.}, vol.~2.\hskip 1em plus 0.5em minus 0.4em\relax IEEE, 2004, pp. II--II.

\bibitem{onishi2008hog}
K.~Onishi, T.~Takiguchi, and Y.~Ariki, ``3d human posture estimation using the hog features from monocular image,'' in \emph{2008 19th International Conference on Pattern Recognition}.\hskip 1em plus 0.5em minus 0.4em\relax IEEE, 2008, pp. 1--4.

\bibitem{pavllo2019semi}
D.~Pavllo, C.~Feichtenhofer, D.~Grangier, and M.~Auli, ``3d human pose estimation in video with temporal convolutions and semi-supervised training,'' in \emph{Proceedings of the IEEE/CVF conference on computer vision and pattern recognition}, 2019, pp. 7753--7762.

\bibitem{liu2020attention}
R.~Liu, J.~Shen, H.~Wang, C.~Chen, S.-c. Cheung, and V.~Asari, ``Attention mechanism exploits temporal contexts: Real-time 3d human pose reconstruction,'' in \emph{Proceedings of the IEEE/CVF Conference on Computer Vision and Pattern Recognition}, 2020, pp. 5064--5073.

\bibitem{chen2021anatomy}
T.~Chen, C.~Fang, X.~Shen, Y.~Zhu, Z.~Chen, and J.~Luo, ``Anatomy-aware 3d human pose estimation with bone-based pose decomposition,'' \emph{IEEE Transactions on Circuits and Systems for Video Technology}, vol.~32, no.~1, pp. 198--209, 2021.

\bibitem{cai2019exploiting}
Y.~Cai, L.~Ge, J.~Liu, J.~Cai, T.-J. Cham, J.~Yuan, and N.~M. Thalmann, ``Exploiting spatial-temporal relationships for 3d pose estimation via graph convolutional networks,'' in \emph{Proceedings of the IEEE/CVF international conference on computer vision}, 2019, pp. 2272--2281.

\bibitem{zhao2023poseformerv2}
Q.~Zhao, C.~Zheng, M.~Liu, P.~Wang, and C.~Chen, ``Poseformerv2: Exploring frequency domain for efficient and robust 3d human pose estimation,'' in \emph{Proceedings of the IEEE/CVF Conference on Computer Vision and Pattern Recognition}, 2023, pp. 8877--8886.

\bibitem{hossain2018lstm}
M.~R.~I. Hossain and J.~J. Little, ``Exploiting temporal information for 3d human pose estimation,'' in \emph{Proceedings of the European conference on computer vision (ECCV)}, 2018, pp. 68--84.

\bibitem{lin2017recurrent}
M.~Lin, L.~Lin, X.~Liang, K.~Wang, and H.~Cheng, ``Recurrent 3d pose sequence machines,'' in \emph{Proceedings of the IEEE conference on computer vision and pattern recognition}, 2017, pp. 810--819.

\bibitem{wang2020motion}
J.~Wang, S.~Yan, Y.~Xiong, and D.~Lin, ``Motion guided 3d pose estimation from videos,'' in \emph{European Conference on Computer Vision}.\hskip 1em plus 0.5em minus 0.4em\relax Springer, 2020, pp. 764--780.

\bibitem{hu2021conditionalgraph}
W.~Hu, C.~Zhang, F.~Zhan, L.~Zhang, and T.-T. Wong, ``Conditional directed graph convolution for 3d human pose estimation,'' in \emph{Proceedings of the 29th ACM International Conference on Multimedia}, 2021, pp. 602--611.

\bibitem{zhang2022mixste}
J.~Zhang, Z.~Tu, J.~Yang, Y.~Chen, and J.~Yuan, ``Mixste: Seq2seq mixed spatio-temporal encoder for 3d human pose estimation in video,'' in \emph{Proceedings of the IEEE/CVF conference on computer vision and pattern recognition}, 2022, pp. 13\,232--13\,242.

\bibitem{tang2023stc}
Z.~Tang, Z.~Qiu, Y.~Hao, R.~Hong, and T.~Yao, ``3d human pose estimation with spatio-temporal criss-cross attention,'' in \emph{Proceedings of the IEEE/CVF Conference on Computer Vision and Pattern Recognition}, 2023, pp. 4790--4799.

\bibitem{choi2022diffupose}
J.~Choi, D.~Shim, and H.~J. Kim, ``Diffupose: Monocular 3d human pose estimation via denoising diffusion probabilistic model,'' \emph{arXiv preprint arXiv:2212.02796}, 2022.

\bibitem{martinez2017simple}
J.~Martinez, R.~Hossain, J.~Romero, and J.~J. Little, ``A simple yet effective baseline for 3d human pose estimation,'' in \emph{Proceedings of the IEEE international conference on computer vision}, 2017, pp. 2640--2649.

\bibitem{pavlakos2017coarse}
G.~Pavlakos, X.~Zhou, K.~G. Derpanis, and K.~Daniilidis, ``Coarse-to-fine volumetric prediction for single-image 3d human pose,'' in \emph{Proceedings of the IEEE conference on computer vision and pattern recognition}, 2017, pp. 7025--7034.

\bibitem{tekin20172dcues}
B.~Tekin, P.~M{\'a}rquez-Neila, M.~Salzmann, and P.~Fua, ``Learning to fuse 2d and 3d image cues for monocular body pose estimation,'' in \emph{Proceedings of the IEEE international conference on computer vision}, 2017, pp. 3941--3950.

\bibitem{pavlakos2018ordinal}
G.~Pavlakos, X.~Zhou, and K.~Daniilidis, ``Ordinal depth supervision for 3d human pose estimation,'' in \emph{Proceedings of the IEEE conference on computer vision and pattern recognition}, 2018, pp. 7307--7316.

\bibitem{yang2023xinxing}
B.~Yang, J.~Huang, X.~Chen, X.~Li, and Y.~Hasegawa, ``Natural grasp intention recognition based on gaze in human--robot interaction,'' \emph{IEEE Journal of Biomedical and Health Informatics}, vol.~27, no.~4, pp. 2059--2070, 2023.

\bibitem{rakotosaona2020pointcleannet}
M.-J. Rakotosaona, V.~La~Barbera, P.~Guerrero, N.~J. Mitra, and M.~Ovsjanikov, ``Pointcleannet: Learning to denoise and remove outliers from dense point clouds,'' in \emph{Computer graphics forum}, vol.~39, no.~1.\hskip 1em plus 0.5em minus 0.4em\relax Wiley Online Library, 2020, pp. 185--203.

\bibitem{wang2019learning}
C.~Wang, H.~Qiu, A.~L. Yuille, and W.~Zeng, ``Learning basis representation to refine 3d human pose estimations,'' in \emph{Proceedings of the AAAI Conference on Artificial intelligence}, vol.~33, no.~01, 2019, pp. 8925--8932.

\bibitem{mei2019learning}
J.~Mei, X.~Chen, C.~Wang, A.~Yuille, X.~Lan, and W.~Zeng, ``Learning to refine 3d human pose sequences,'' in \emph{2019 International Conference on 3D Vision (3DV)}.\hskip 1em plus 0.5em minus 0.4em\relax IEEE, 2019, pp. 358--366.

\bibitem{zeng2022smoothnet}
A.~Zeng, L.~Yang, X.~Ju, J.~Li, J.~Wang, and Q.~Xu, ``Smoothnet: A plug-and-play network for refining human poses in videos,'' in \emph{European Conference on Computer Vision}.\hskip 1em plus 0.5em minus 0.4em\relax Springer, 2022, pp. 625--642.

\bibitem{lstm}
S.~Hochreiter and J.~Schmidhuber, ``Long short-term memory,'' \emph{Neural computation}, vol.~9, no.~8, pp. 1735--1780, 1997.

\bibitem{gcn}
T.~N. Kipf and M.~Welling, ``Semi-supervised classification with graph convolutional networks,'' \emph{arXiv preprint arXiv:1609.02907}, 2016.

\bibitem{moon2019posefix}
G.~Moon, J.~Y. Chang, and K.~M. Lee, ``Posefix: Model-agnostic general human pose refinement network,'' in \emph{Proceedings of the IEEE/CVF Conference on Computer Vision and Pattern Recognition}, 2019, pp. 7773--7781.

\bibitem{wang2020graph}
J.~Wang, X.~Long, Y.~Gao, E.~Ding, and S.~Wen, ``Graph-pcnn: Two stage human pose estimation with graph pose refinement,'' in \emph{Computer Vision--ECCV 2020: 16th European Conference, Glasgow, UK, August 23--28, 2020, Proceedings, Part XI 16}.\hskip 1em plus 0.5em minus 0.4em\relax Springer, 2020, pp. 492--508.

\bibitem{ho2020ddpm}
J.~Ho, A.~Jain, and P.~Abbeel, ``Denoising diffusion probabilistic models,'' \emph{Advances in neural information processing systems}, vol.~33, pp. 6840--6851, 2020.

\bibitem{saharia2022superres}
C.~Saharia, J.~Ho, W.~Chan, T.~Salimans, D.~J. Fleet, and M.~Norouzi, ``Image super-resolution via iterative refinement,'' \emph{IEEE Transactions on Pattern Analysis and Machine Intelligence}, vol.~45, no.~4, pp. 4713--4726, 2022.

\bibitem{meng2021sdedit}
C.~Meng, Y.~He, Y.~Song, J.~Song, J.~Wu, J.-Y. Zhu, and S.~Ermon, ``Sdedit: Guided image synthesis and editing with stochastic differential equations,'' \emph{arXiv preprint arXiv:2108.01073}, 2021.

\bibitem{shan2023diffusion}
W.~Shan, Z.~Liu, X.~Zhang, Z.~Wang, K.~Han, S.~Wang, S.~Ma, and W.~Gao, ``Diffusion-based 3d human pose estimation with multi-hypothesis aggregation,'' \emph{arXiv preprint arXiv:2303.11579}, 2023.

\bibitem{gong2023diffpose}
J.~Gong, L.~G. Foo, Z.~Fan, Q.~Ke, H.~Rahmani, and J.~Liu, ``Diffpose: Toward more reliable 3d pose estimation,'' in \emph{Proceedings of the IEEE/CVF Conference on Computer Vision and Pattern Recognition}, 2023, pp. 13\,041--13\,051.

\bibitem{zhou2023diff3dhpe}
J.~Zhou, T.~Zhang, Z.~Hayder, L.~Petersson, and M.~Harandi, ``Diff3dhpe: A diffusion model for 3d human pose estimation,'' in \emph{Proceedings of the IEEE/CVF International Conference on Computer Vision}, 2023, pp. 2092--2102.

\bibitem{zhao2022graformer}
W.~Zhao, W.~Wang, and Y.~Tian, ``Graformer: Graph-oriented transformer for 3d pose estimation,'' in \emph{Proceedings of the IEEE/CVF Conference on Computer Vision and Pattern Recognition}, 2022, pp. 20\,438--20\,447.

\bibitem{song2020ddim}
J.~Song, C.~Meng, and S.~Ermon, ``Denoising diffusion implicit models,'' \emph{arXiv preprint arXiv:2010.02502}, 2020.

\bibitem{chen2018cpn}
Y.~Chen, Z.~Wang, Y.~Peng, Z.~Zhang, G.~Yu, and J.~Sun, ``Cascaded pyramid network for multi-person pose estimation,'' in \emph{Proceedings of the IEEE conference on computer vision and pattern recognition}, 2018, pp. 7103--7112.

\bibitem{ionescu2013h36m}
C.~Ionescu, D.~Papava, V.~Olaru, and C.~Sminchisescu, ``Human3. 6m: Large scale datasets and predictive methods for 3d human sensing in natural environments,'' \emph{IEEE transactions on pattern analysis and machine intelligence}, vol.~36, no.~7, pp. 1325--1339, 2013.

\bibitem{paszke2019pytorch}
A.~Paszke, S.~Gross, F.~Massa, A.~Lerer, J.~Bradbury, G.~Chanan, T.~Killeen, Z.~Lin, N.~Gimelshein, L.~Antiga \emph{et~al.}, ``Pytorch: An imperative style, high-performance deep learning library,'' \emph{Advances in neural information processing systems}, vol.~32, 2019.

\bibitem{nichol2021cosine}
A.~Q. Nichol and P.~Dhariwal, ``Improved denoising diffusion probabilistic models,'' in \emph{International Conference on Machine Learning}.\hskip 1em plus 0.5em minus 0.4em\relax PMLR, 2021, pp. 8162--8171.

\bibitem{kingma2014adam}
D.~P. Kingma and J.~Ba, ``Adam: A method for stochastic optimization,'' \emph{arXiv preprint arXiv:1412.6980}, 2014.

\bibitem{shan2022tds}
W.~Shan, Z.~Liu, X.~Zhang, S.~Wang, S.~Ma, and W.~Gao, ``P-stmo: Pre-trained spatial temporal many-to-one model for 3d human pose estimation,'' in \emph{European Conference on Computer Vision}.\hskip 1em plus 0.5em minus 0.4em\relax Springer, 2022, pp. 461--478.

\end{thebibliography}

\end{document}